\documentclass{article}

\usepackage{arxiv}

\usepackage[T1]{fontenc}
\usepackage{microtype}

\usepackage{amsmath, amssymb, mathtools}

\usepackage{graphicx}
\graphicspath{{figures/}}
\usepackage{subcaption}
\usepackage{booktabs}
\usepackage{wrapfig}

\usepackage{tikz}
\usetikzlibrary{arrows.meta, backgrounds, calc, decorations.pathreplacing, fit}

\definecolor{pfink}{HTML}{32312D}
\definecolor{pfmuted}{HTML}{6E6A63}
\definecolor{pffaint}{HTML}{B9B4A9}
\definecolor{pflat}{HTML}{9FBBD6}
\definecolor{pfobs}{HTML}{2F5F8F}
\definecolor{pfaccent}{HTML}{C26B5C}
\definecolor{pftokA}{HTML}{DBA94F}
\definecolor{pftokB}{HTML}{5A9E8E}
\definecolor{pftokC}{HTML}{9A7CB4}
\definecolor{pftokD}{HTML}{C2687F}

\tikzset{
  pfpicture/.style={
    x=1cm, y=1cm,
    every node/.append style={font=\sffamily\footnotesize, inner sep=0pt},
  },
  pflbl/.style={font=\sffamily\footnotesize, color=pfink},
  pfsub/.style={font=\sffamily\scriptsize, color=pfmuted},
  pfband/.style={font=\sffamily\small, color=pfink},
  pfhead/.style={font=\sffamily\footnotesize, color=pfink, anchor=north},
  pfblock/.style={rectangle, rounded corners=2.2pt, draw=pffaint,
                  line width=0.5pt, fill=none},
  pflatent/.style={circle, draw=pfink, line width=0.5pt, fill=pflat,
                   minimum size=3.2mm},
  pfobserved/.style={circle, draw=pfink, line width=0.5pt, fill=pfobs,
                     minimum size=3.2mm},
  pfmarked/.style={circle, draw=pfaccent, line width=1.1pt, fill=pflat,
                   minimum size=3.2mm},
  pfwire/.style={line width=0.45pt, color=pfink, opacity=0.45,
                 -{Stealth[length=2.1pt, width=2.1pt]},
                 shorten >=0.4pt, shorten <=0.4pt},
  pfwiremark/.style={line width=0.9pt, color=pfaccent, opacity=0.95,
                     -{Stealth[length=2.7pt, width=2.7pt]},
                     shorten >=0.5pt, shorten <=0.5pt},
  pfarrow/.style={-{Stealth[length=2.7pt, width=2.7pt]}, line width=0.5pt,
                  color=pfink},
  pfop/.style={rectangle, rounded corners=0.9pt, draw=pfink, line width=0.5pt,
               fill=white, minimum height=4.0mm, minimum width=5.2mm,
               inner sep=1.5pt, font=\sffamily\footnotesize},
  pfqbox/.style={rectangle, rounded corners=0.8pt, draw=pfink,
                 line width=0.45pt, fill=white, minimum width=4.0mm,
                 minimum height=3.4mm, inner sep=1pt, font=\sffamily\scriptsize},
  pftoken/.style={rectangle, rounded corners=0.4pt, draw=pfink,
                  line width=0.4pt, minimum width=3.9mm, minimum height=3.5mm,
                  font=\sffamily\scriptsize},
}

\usepackage{acro}

\makeatletter
\@ifpackageloaded{natbib}{}{\usepackage[numbers,sort&compress]{natbib}}
\makeatother

\usepackage{hyperref}
\usepackage[capitalize, noabbrev]{cleveref}

\newcommand{\pflm}{PFLM}

\DeclareAcronym{scm}{short = SCM, long = structural causal model}
\DeclareAcronym{hmm}{short = HMM, long = hidden Markov model}
\DeclareAcronym{utm}{short = UTM, long = universal Turing machine}
\DeclareAcronym{icl}{short = ICL, long = in-context learning}

\DeclareMathOperator{\clip}{clip}
\DeclareMathOperator{\pa}{pa}

\newif\ifpfanonymous
\newcommand{\codeurl}{\url{https://github.com/cbl/prior-fitted-language-model}}
\newcommand{\weightsurl}{\url{https://hf.co/lennartcb/pflm1}}
\newcommand{\releasefootnote}{\ifpfanonymous\else\footnote{Code: \codeurl{}. Weights: \weightsurl{}.}\fi}

\title{Learning to Learn a Language}
\author{%
  Lennart Carstens-Behrens \\
  Fraunhofer Institute for Algorithms and Scientific Computing SCAI \\
  University Hospital Bonn \\
  \texttt{lennart.carstens-behrens@scai.fraunhofer.de}
  \And
  Holger Fr\"ohlich \\
  Fraunhofer Institute for Algorithms and Scientific Computing SCAI \\
  University Hospital Bonn \\
  \texttt{holger.froehlich@scai.fraunhofer.de}%
}
\date{\today}

\begin{document}
\maketitle

\begin{abstract}
We present the Prior-Fitted Language Model (\pflm{}), a 300M-parameter byte-level transformer pretrained only on samples from a synthetic non-linguistic prior. Given a prefix of real text, it learns to predict the language in context with frozen weights, having never seen a word of any real language. Every training sequence is generated by a recurrent structural causal model drawn fresh from a distribution over such models. The model never sees the same language twice during training, so the only way to predict the continuation is to infer the language from the prefix. Samples from this prior share the statistical signatures of natural text: Zipfian frequencies, slow entropy-rate convergence, and long-range dependence. On Wikipedia in six languages, bits per byte fall from the uniform eight to between $0.9$ and $2.4$ at one million bytes of context. Given numerals instead of text, \pflm{} learns to count, to compare magnitudes, and to add approximately. It predicts deterministic sequences like Rudin--Shapiro or the prime indicator, and it compresses six non-text domains, from source code to speech, below gzip and PPMd. The model has not learned a language. It has learned to learn one.

\end{abstract}

\begin{figure}[h!]
  \centering
  \begin{subfigure}[t]{0.49\linewidth}
    \includegraphics[width=\linewidth]{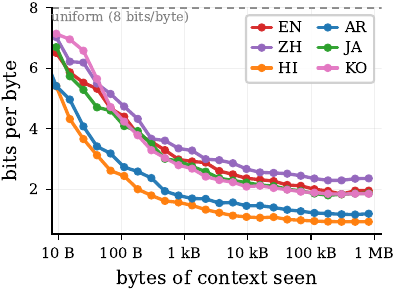}
    \caption{Natural language.}
    \label{fig:headline:language}
  \end{subfigure}\hfill
  \begin{subfigure}[t]{0.49\linewidth}
    \includegraphics[width=\linewidth]{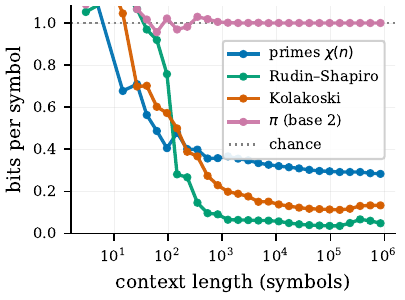}
    \caption{Deterministic sequences.}
    \label{fig:headline:sequences}
  \end{subfigure}
  \caption{%
    \textbf{PFLM, trained only on the synthetic non-linguistic prior of \cref{sec:method}, learns to predict real languages and deterministic sequences in context with frozen weights.}
    \textbf{(a)} Bits per byte against bytes of context seen, for English, Chinese, Hindi, Arabic, Japanese, and Korean Wikipedia (UTF-8). Each curve is the mean over $32$ held-out windows of one million bytes, averaged in log-spaced bins. Every curve falls from the uniform rate of eight bits to between $0.9$ and $2.4$ bits at one million bytes. 
    \textbf{(b)} Bits per symbol against symbols seen for four binary sequences. By $10^6$ symbols Rudin--Shapiro costs $0.04$ bits, Kolakoski $0.13$, and the prime indicator $0.28$, against $0.37$ for a predictor that knows only the density of primes at that position; $\pi$ stays at $1.00$ throughout.
  }
  \label{fig:headline}
\end{figure}

\clearpage
\section{Introduction}
\label{sec:intro}

What evolution gives a child is not a language but the ability to acquire one. After tens of millions of words, a child is fluent in whichever language surrounds them. A modern language model is trained on tens of trillions. This six-order-of-magnitude gap is not a fact about language. It is a fact about how much prior knowledge each brings to the data.

The child's prior is not knowledge of any specific language but expectations about what any language can be: wide enough to admit any of them, narrow enough to learn one. Human languages occupy a bounded region in the space of possible symbol systems, and the child's hypothesis space is that region. A language model begins with no such prior. Its inductive biases are architectural: generic pressures toward smoothness and compositionality that say nothing about what a language can be. Every regularity it learns about language must be inferred from the corpus alone.

This is why models lag at inferences humans do effortlessly. A baby learns whichever language surrounds them from the speech they hear. A child, born with a sense of quantity, gradually learns to count, compare, and do arithmetic. A mathematician recognizes the rule producing an unfamiliar sequence from its first terms. Each rests on prior knowledge of what languages, what numbers, or what sequences can be: evolution and experience supply the structure on which finite evidence rests. The model has neither, and must extract from trillions of tokens what humans inherit or learn.

The asymmetry suggests a recipe. Equip the model with a prior, not over one language but over the family of possible languages. Sample a fresh language for every pretraining sequence; the model never sees the same one twice. It cannot memorize any specific language, only learn to acquire one from a prefix. The pretraining task is meta-learning by construction. The trained network performs amortized Bayesian inference over the prior. \emph{Prior-Fitted Networks} \citep{mueller2022,hollmann2023} establish the recipe on tabular classification and regression, since extended to various settings such as in-context regression of function classes \citep{garg2022}, survival analysis \citep{seletkov2026}, and algorithmic sequences from random programs \citep{graumoya2024}. We carry it to language: a prior over synthetic ``languages'' (where we use the term ``language'' to indicate the domain of structured sequences) whose samples pre-train a transformer, the \emph{Prior-Fitted Language Model} (\pflm{}).\releasefootnote{} To the best of our knowledge, no prior work has shown a language model learning to predict natural language in context after pretraining only on samples from a constructed non-linguistic prior. Every language model to date, however synthetic its curriculum, has been trained on natural language \citep{eldan2023,gunasekar2023,mccoy2025}.

Our prior is a recurrent \ac{scm}. One independently sampled \ac{scm} defines one synthetic language. The \ac{scm} evolves latent state through a random graph of structural equations, emitting one token per step. Each token depends, by construction, on the tokens before it. The prior's samples reproduce the signature statistics of natural text: Zipfian token frequencies, slow entropy-rate convergence, mutual-information decay at multiple scales. The prior is constructed, not fit; its samples occupy the same statistical neighborhood as natural text without sharing any specific script.

A byte-level transformer trained on samples from this prior infers natural language in context. Its training data contains no natural-language text. On every language we evaluate, bits per byte falls steadily with increasing context (\cref{fig:headline}), from the uniform rate of eight to between $0.9$ and $2.4$ at one million bytes.

Given numerals in context, the same model learns to count, to compare quantities, and to add approximately, the pattern of number sense seen in children before they can calculate \citep{dehaene1997,barth2005}. Beyond language and number, the model predicts deterministic sequences like Rudin--Shapiro~\citep{allouche2003}, Kolakoski~\citep{kolakoski1965}, or the prime indicator, where each term follows from a generating rule rather than a frequency.

\section{The Synthetic-Language Prior}
\label{sec:method}

\pflm{} is trained on synthetic data alone: every pretraining sequence is a sample from a prior $\pi$ over synthetic \emph{languages}.
This section instantiates the idea (\cref{sec:method:inference}), builds the prior (\cref{sec:method:language,sec:method:prior}), and specifies the model that approximates the predictor it induces (\cref{sec:method:model}).

\subsection{In-context inference over languages}
\label{sec:method:inference}

A fresh language generates every training sequence, and the model sees only the resulting tokens, never the language that produced them.
The only way to drive down next-token loss is to infer the generating language from the prefix.
Next-token prediction is thus Bayesian inference \citep{xie2022}.
The predictor that minimizes the pretraining loss is the posterior predictive over languages,
\begin{equation}
  p^\star(y_t \mid y_{<t})
  \;=\;
  \int p(y_t \mid y_{<t}, \ell)\; p(\ell \mid y_{<t}) \,\mathrm{d}\ell,
  \qquad
  p(\ell \mid y_{<t}) \;\propto\; \pi(\ell)\, p(y_{<t} \mid \ell),
  \label{eq:posterior}
\end{equation}
which weights each language by how well it explains the tokens seen so far, then averages their next-token predictions under that weighting.
As the prefix lengthens, the posterior $p(\ell \mid y_{<t})$ concentrates on the generating language and the prediction sharpens.

\subsection{A synthetic language as a recurrent SCM}
\label{sec:method:language}

\begin{figure}[t]
  \centering
  \resizebox{\linewidth}{!}{\input{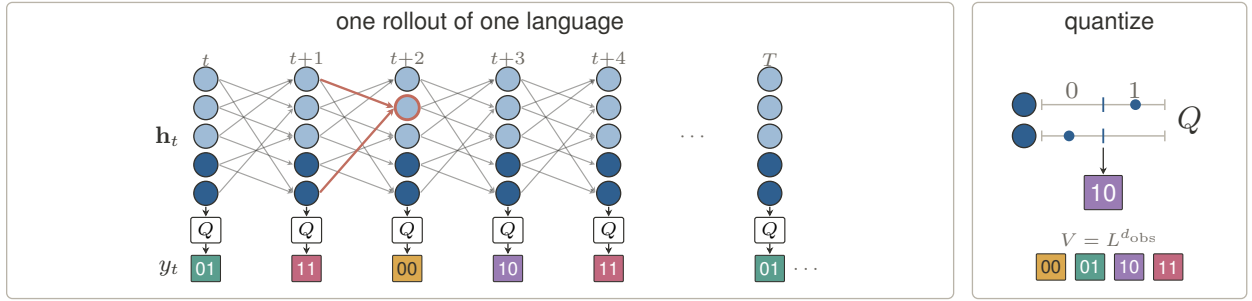}}
  \caption{%
    \textbf{Every pretraining sequence is generated by a small recurrent program drawn from the prior.}
    \emph{(a)} A synthetic language $\ell$ is a sparse recurrent \ac{scm}. Its wiring, each node's structural equation $f_i$ and activation $\sigma_i$, and the noise scale $\eta$ are all sampled, so no two sequences share a language. One node is marked and its update shown in full (\cref{eq:dynamics}); the tiles are single draws from the six equation families of \cref{tab:equations}, each plotted over the plane of two of its parents and clipped to $[-1,1]$ as the dynamics does.
    \emph{(b)} Rolling that language forward emits one token per step, by quantizing the $d_{\mathrm{obs}}$ observed coordinates and reading the resulting digits as one symbol (\cref{eq:emission}). The drawing uses $L=2$ and $d_{\mathrm{obs}}=2$ for legibility; the paper uses $L=2$ and $d_{\mathrm{obs}}=8$, so the token alphabet is exactly the $256$ byte values.
    Only the tokens leave the generator. Because $\ell$ is never revealed and never drawn twice, the only way to predict the next token is to infer the language from those already seen.%
  }
  \label{fig:prior}
\end{figure}

We draw each pretraining sequence from an independently sampled recurrent \ac{scm} \citep{pearl2009}, which we call a \emph{synthetic language}.

A synthetic language is a small directed graph: a few dozen latent nodes sparsely connected by a randomly drawn pattern of edges.
Each node carries a continuous latent state that evolves over time as illustrated in \cref{fig:prior}(b).
At each timestep, every node updates from the states of its parents, and the state of the first few nodes is then quantized into a single output token.
Both the wiring of the graph and the update rules are drawn fresh for every language. 

Formally, a synthetic language is the tuple
\[
  \ell \;=\; \big(G,\, \{f_i\},\, \{\sigma_i\},\, \{\rho_i\},\, \eta,\, Q\big).
\]
The directed graph $G$ on $d$ latent nodes carries three per-node components.
At node $i$, $f_i$ is a structural equation and $\sigma_i$ a scalar activation.
The firing-rate function $\rho_i \colon \mathbb{N} \to \{0, 1\}$ takes value $1$ on the steps at which the node updates and $0$ on the steps it skips.
A single noise scale $\eta \geq 0$ governs every node, and a single quantizer $Q$ maps the continuous state to discrete tokens.
The graph specifies a parent set $\pa(i) \subseteq \{1, \ldots, d\}$ for each node, with self-loops permitted.

The latent state $\mathbf{h}_t \in [-1, 1]^d$ evolves by a noisy structural update gated by the firing-rate function,
\begin{equation}
  h_{t+1,\,i}
  \;=\;
  \begin{cases}
    \clip\!\big(\, \sigma_i\big( f_i(\mathbf{h}_{t,\,\pa(i)}) \big) + \epsilon_{t,\,i} \,\big) & \text{if } \rho_i(t) = 1, \\
    h_{t,\,i} & \text{otherwise,}
  \end{cases}
  \label{eq:dynamics}
\end{equation}
where $\mathbf{h}_{t,\,\pa(i)}$ collects the parent values, $\clip(x) = \max(-1, \min(1, x))$ keeps the state bounded, and the noise $\epsilon_{t,\,i} \sim \mathcal{N}(0, \eta^2)$ is independent across nodes and timesteps.

A timestep composes $K \geq 1$ such updates, each with its own parent sets, equations, and activations, applied sequentially like the layers of a feed-forward network.
Every layer reads the state left by the previous one and writes all its nodes in parallel, and the token is emitted after the state has propagated through all $K$ layers.

A node with $\rho_i \equiv 1$ updates every step, as in a standard recurrent \ac{scm}; a node with rate four updates every fourth step and holds its value in between.
Mixing rates across nodes lets one rolled-out language carry dependencies at multiple timescales, reproducing the long-range structure of natural language (\cref{fig:mi_decay}).
We use periodic schedules with power-of-two rates as a simple family that spans a geometric ladder of timescales, from nodes that change every step to a thin tail that changes every $2, 4, 8, \ldots$ steps.

A token is emitted at each step from the first $d_{\mathrm{obs}} \leq d$ coordinates of the state,
\begin{equation}
  y_t \;=\; Q(h_{t,\,1}, \ldots, h_{t,\,d_{\mathrm{obs}}}) \;\in\; \{0, 1, \ldots, V - 1\}.
  \label{eq:emission}
\end{equation}
The quantizer is finite scalar quantization \citep{mentzer2024}: it bins each coordinate into $L$ uniform intervals in $[-1, 1]$ and combines the bins by mixed-radix encoding, so the vocabulary size is $V = L^{d_{\mathrm{obs}}}$.
The choice $L = 2$, $d_{\mathrm{obs}} = 8$ recovers a vocabulary of raw bytes ($V = 256$), the setting we use throughout the paper.

The noise scale $\eta$ adds the intrinsic unpredictability of natural text and guarantees that every token has positive probability of being emitted at each step. Hence $\pi$ assigns positive probability to every finite token sequence, and the posterior predictive \cref{eq:posterior} is well defined for arbitrary prefixes, not only those typical of the prior.

The difficulty of prediction lives in the emitted sequence of discrete tokens. The latent state is first-order Markov: given $\mathbf{h}_t$, the past is irrelevant. The observed process is not. Each token $y_t$ quantizes the first $d_{\mathrm{obs}}$ coordinates and discards the rest, a lossy view of the Markov chain. The unobserved coordinates act as hidden memory, so the optimal predictor of $y_{t+1}$ is a filter, maintaining a posterior over $\mathbf{h}_t$ given the full prefix.

\subsection{The prior over languages}
\label{sec:method:prior}

We sample $\ell \sim \pi$ by drawing each component of the tuple independently.
The graph is built node by node: each node receives a parent set whose size is sampled uniformly from a per-language range, with members drawn without replacement from $\{1, \ldots, d\}$.
Each structural equation $f_i$ is drawn per node from one of six functional families (\cref{tab:equations}), with coefficients from Gaussians whose widths are sampled per language.
Each activation $\sigma_i$ is drawn from $\{\tanh, \mathrm{softsign}, \clip, \sin, \mathrm{id}\}$, the noise scale $\eta$ from a fixed range, and the depth $K$ from a per-language range.
Each $\rho_i$ is periodic: node $i$ updates at the steps $t$ with $t \equiv \phi_i \pmod{r_i}$ and skips the others.
The rate $r_i \in \{1, 2, 4, \ldots\}$ is drawn from a geometric-style distribution over powers of two, biased so most nodes fire every step and a small tail updates every 2, 4, 8, or more.
The phase $\phi_i$ is drawn uniformly from $\{0, \ldots, r_i - 1\}$, which de-synchronizes nodes with the same rate.
The hyperparameter ranges are listed in \cref{app:prior-config}.

\begin{table}[t]
  \centering
  \caption{%
    \textbf{The six structural-equation families.}
    Each node draws its update $f_i$ from one of these, applied to its parent values $\mathbf{p} \equiv \mathbf{h}_{t,\,\pa(i)}$.
    Coefficients are sampled from per-language Gaussians, and the structural sizes (MLP hidden width, Fourier feature count, tree depth) are drawn per language as well.%
  }
  \label{tab:equations}
  \begin{tabular}{lll}
    \toprule
    \textbf{Family} & \textbf{Form} & \textbf{Parameters} \\
    \midrule
    Linear map       & $\mathbf{w}^\top \mathbf{p} + b$                                                  & $\mathbf{w},\, b$ \\
    Polynomial       & $b + \sum_i \sum_{k=1}^{3} a_{ki}\, p_i^k + \sum_{i<j} c_{ij}\, p_i p_j$           & $a,\, c,\, b$ \\
    Fourier features & $b + \sqrt{2}\, \sum_j w_j \cos(\boldsymbol{\omega}_j^\top \mathbf{p} + \varphi_j)$ & $\boldsymbol{\omega}_j,\, \varphi_j,\, w_j,\, b$ \\
    MLP (two-layer)  & $\mathbf{w}_2^\top \tanh(W_1 \mathbf{p} + \mathbf{b}_1) + b_2$                     & $W_1,\, \mathbf{b}_1,\, \mathbf{w}_2,\, b_2$ \\
    Decision tree    & threshold tests on $\mathbf{p}$ route to a leaf value                             & splits, thresholds, leaves \\
    Step function    & $v_{\,\mathrm{bin}(\mathbf{w}^\top \mathbf{p} + b)}$, four bins                    & $\mathbf{w},\, b$, edges, levels \\
    \bottomrule
  \end{tabular}
\end{table}

To produce one pretraining sequence we draw a language $\ell \sim \pi$, initialize $\mathbf{h}_0 \in [-1, 1]^d$, roll the dynamics in \cref{eq:dynamics} forward for $T$ steps, and emit the token stream $(y_1, \ldots, y_T)$ via \cref{eq:emission}, one token per step as drawn in \cref{fig:prior}(b).
The initialization law is itself sampled per sequence from six families on $[-1, 1]$: uniform on a sub-interval, Gaussian, Beta, two-point near $\pm 1$, bimodal, and sparse, with most coordinates exactly zero.

\subsection{Model and training}
\label{sec:method:model}

\pflm{} is a 300M-parameter decoder-only transformer (\cref{tab:model-config}) trained by next-token prediction to approximate the posterior predictive of \cref{eq:posterior}.
The model is a hybrid stack similar to Qwen3-Next \citep{qwen2025}: gated linear-attention layers \citep{yang2025} interleaved with attention layers in a 3:1 ratio, the attention restricted to a sliding window of $2^{18}$ tokens.

The linear-attention layers run in linear time and carry a recurrent state across the sequence, while the sliding-window attention layers add exact recall within that window, so the stack is sub-quadratic and trains on sequences far longer than full attention permits.

The vocabulary is the $256$ byte values, the emission alphabet of the prior (\cref{eq:emission}): one token is one byte, so synthetic streams and UTF-8 text are sequences over the same alphabet.
We train \pflm{} on $150$ billion tokens drawn afresh from $\pi$ (\cref{sec:method:prior}) in six stages of increasing context length, $2^{10}$ to $2^{20}$ tokens, for $20$, $60$, $40$, $15$, $10$, and $5$ billion tokens respectively, each stage initialized from the last, with the NorMuon optimizer \citep{li2025} and a warmup-stable-decay learning-rate schedule \citep{hu2024}.
\Cref{app:training} reports the remaining hyperparameters.

\section{Statistical Signatures of the Prior}
\label{sec:signatures}

Samples from the \ac{scm} prior match natural language across its statistical signatures, from byte marginals to long-range dependencies.
We test against four data-free comparators whose parameters are sampled rather than fitted on text: i.i.d. uniform, i.i.d. Zipf, a random bigram, and a random \ac{hmm}.
The natural-language sources are English, Chinese, Hindi, Arabic, Japanese, and Korean Wikipedia, each contributing 500 sequences of 4096 bytes.
The prior is sampled here at the configuration of the family that best matches these statistics. The training configuration (\cref{app:prior-config}) is instead chosen on transfer to natural language.

\subsection{Scalar fingerprints}
\label{sec:signatures:scalar}

\begin{figure}[t]
  \centering
  \includegraphics[width=\linewidth]{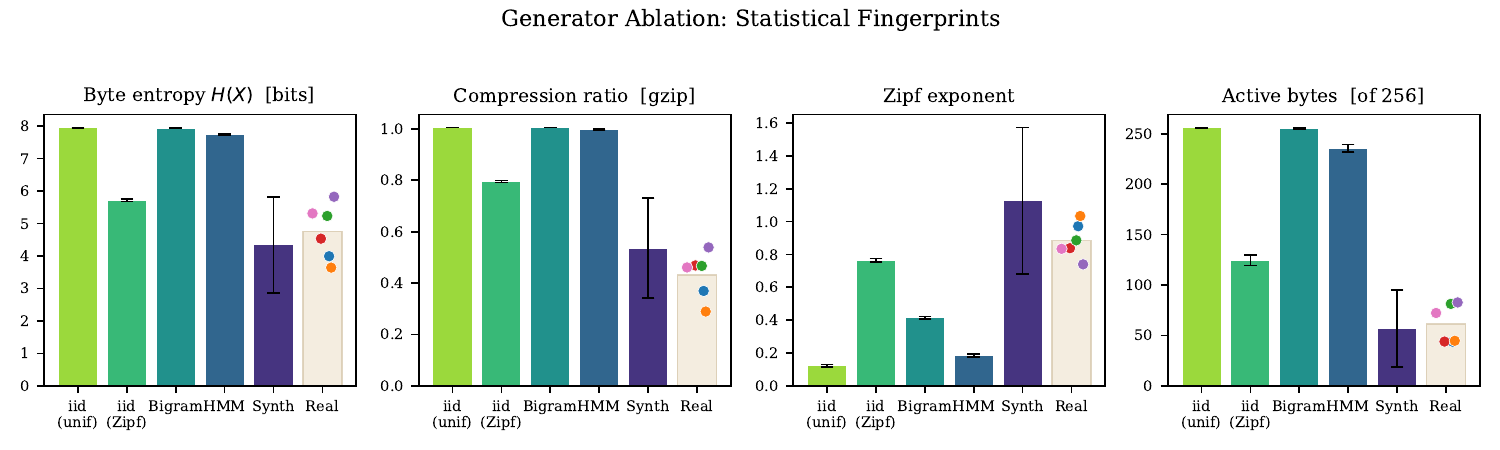}
  \caption{%
    \textbf{The \ac{scm} prior matches natural language on every scalar fingerprint; classical generators each fail on at least one.}
    For each source, mean $\pm$ standard deviation across 500 sequences of 4096 bytes.
    Panels (left to right): byte entropy, written $H(X)$ in the figure, gzip compression ratio, bigram Zipf exponent, active byte count.
    ``Synth'' is the \ac{scm} prior; ``Real'' shows the mean across the six natural languages with individual values as colored dots.
    The i.i.d. uniform, bigram, and \ac{hmm} baselines produce near-uniform marginals (entropy $\approx 8$, gzip $\approx 1$, all 256 bytes active);
    i.i.d. Zipf corrects the marginal but not the conditional structure;
    only the \ac{scm} prior's range overlaps the natural-language values on every panel.%
  }
  \label{fig:staircase}
\end{figure}

Each baseline fails differently.
The i.i.d. uniform, bigram, and \ac{hmm} baselines produce near-uniform marginal distributions: byte entropy at the $\log_2 256 = 8$-bit ceiling, gzip ratio near one, all 256 bytes active.
The i.i.d. Zipf baseline fixes the marginal \citep{zipf1949,piantadosi2014} but lacks conditional structure between adjacent bytes, missing the natural-language cluster on compression and Zipf exponent.
Only the \ac{scm} prior lies in the natural-language band on all four statistics.

\subsection{Trajectory statistics}
\label{sec:signatures:trajectory}

\begin{figure}[t]
  \centering
  \begin{subfigure}[t]{0.49\linewidth}
    \includegraphics[width=\linewidth]{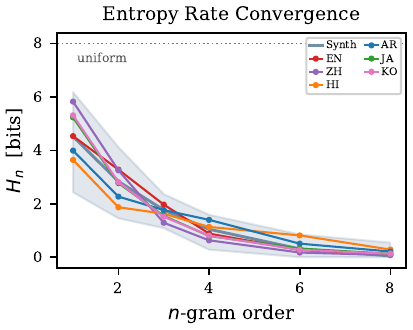}
    \caption{Entropy rate convergence.}
    \label{fig:entropy_rate}
  \end{subfigure}\hfill
  \begin{subfigure}[t]{0.49\linewidth}
    \includegraphics[width=\linewidth]{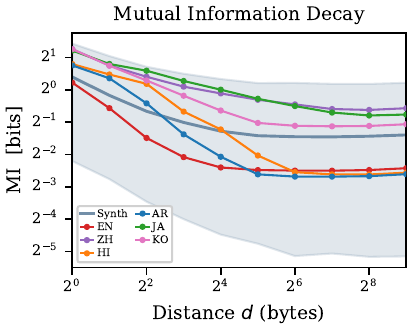}
    \caption{Mutual information decay.}
    \label{fig:mi_decay}
  \end{subfigure}
  \caption{%
    \textbf{The \ac{scm} prior reproduces both the slow entropy-rate convergence and the slow long-range MI decay of natural language.}
    Synth shows the median (line) and 10--90 percentile range (band) across 500 \ac{scm} samples;
    each natural language is the per-sequence median across 500 wiki sequences.
    \textbf{(a)} Conditional entropy $H(Y_t \mid Y_{t-1}, \ldots, Y_{t-n+1})$ at $n$-gram orders 1--8.
    The synth band covers every natural-language curve at every order;
    dotted reference at $H = \log_2 256 = 8$ bits is the i.i.d. uniform-random ceiling.
    \textbf{(b)} $\mathrm{MI}(Y_t;\, Y_{t+d})$ at log-spaced distances $d = 1, 2, 4, \ldots, 512$ bytes.
    Synth and natural decay together from $d=1$ to $d=512$ bytes.%
  }
  \label{fig:trajectory}
\end{figure}

Natural languages converge their entropy rate slowly \citep{shannon1951,brown1992}: each successive order of context strips less than a bit from the conditional entropy (\cref{fig:entropy_rate}).
This reflects weak but persistent dependencies at every scale.
The \ac{scm} prior produces the same trajectory: its median tracks the natural-language band from $\sim 5$ bits at $n=1$ down to under one bit at $n=8$.
The 10--90 envelope contains every natural-language curve.

Long-range structure is the most discriminating signature \citep{ebeling1994,debowski2015} (\cref{fig:mi_decay}).
Natural-language MI decays slowly without dropping to zero, from $\sim 2$ bits at $d=1$ to $0.1$--$0.5$ bits at $d=512$ \citep{lin2017}.
The \ac{scm} prior's median follows the same shape, and its 10--90 envelope brackets every natural-language curve.
This long-range MI traces specifically to the multi-rate firing schedule.
An ablation that fixes $\rho_i \equiv 1$ for every node leaves the scalar fingerprints unchanged but reduces the prior's long-range mutual information, integrated over distance, by 18\%; the tail of slow nodes is what carries dependencies across many steps.

\section{In-Context Learning on Real Data}
\label{sec:experiments}

Every result in this section is produced from a single model, trained only on samples from the synthetic prior of \cref{sec:method}.
We report bits per byte (BPB) for the byte streams, accuracy for the number tasks, and bits per symbol for the digit sequences (each over a binary alphabet, so chance is one bit). \Cref{app:eval} gives the corpora and protocol.

\subsection{Natural language}
\label{sec:exp:language}

On six languages, the model lowers its cost as the context grows, from the uniform rate of eight bits to between $0.9$ and $2.4$ bits at one million bytes (\cref{fig:headline:language}).
Each curve conditions the model on a growing prefix of Wikipedia, in English, Chinese, Hindi, Arabic, Japanese, and Korean, and measures bits per byte at each position.

The curves differ in level for two reasons.
Under UTF-8 a character costs one byte in the Latin script, two in Arabic, and three in Devanagari and in the Chinese, Japanese, and Korean scripts, so bits per byte is a different fraction of bits per character for each script (\cref{app:eval}).
The remaining difference is the predictability of the language and of the corpus.

\subsection{Beyond language}
\label{sec:exp:beyond}

\begin{figure}[t]
  \centering
  \includegraphics[width=\linewidth]{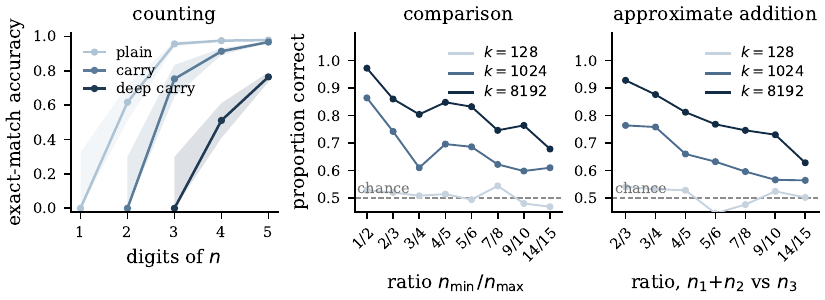}
  \caption{%
    \textbf{Given numerals in context, PFLM learns to count through carries, to order three-digit numerals by magnitude, and to judge approximate sums.}
    \emph{Left:} exact match of the next numeral while counting from $2$ to $99{,}999$, by number of digits, for increments with no carry, one carry, and a carry across two or more trailing nines; bands are 95\% Wilson intervals.
    \emph{Middle:} accuracy at naming the larger of two three-digit numerals, by the ratio of smaller to larger, after $k$ in-context examples; at $8192$ examples accuracy falls from $0.97$ at ratio $1/2$ to $0.68$ at $14/15$.
    \emph{Right:} accuracy at judging whether $n_1 + n_2$ exceeds $n_3$, by the ratio of the sum to $n_3$; $n_3$ exceeds either addend, so comparing one addend to $n_3$ scores at chance \citep{barth2005}.
    Chance is one half; $500$ queries per ratio bin and shot count, none of them among the examples.
  }
  \label{fig:number}
\end{figure}

\paragraph{Number.}
We test three abilities on numerals in context: counting, magnitude comparison, and approximate addition (\cref{fig:number}).
Counting upward, the model produces the next numeral exactly by three digits when no carry is needed, one digit later when the increment carries, and two digits later when it carries across trailing nines.
Asked which of two three-digit numerals is larger, accuracy at $8192$ examples falls from $0.97$ when one is half the other to $0.68$ when they differ by a fifteenth, the distance effect Moyer and Landauer observed in people comparing digits \citep{moyer1967}.
Asked whether the sum of two numerals exceeds a third, accuracy falls the same way, from $0.93$ to $0.62$, among trios whose sum and third number have the same digit count, so no length cue carries it.
A similar ratio dependence, the signature of the approximate number sense \citep{dehaene1997}, can be observed in preschool children, who learn to add and compare quantities before they learn to compute exact sums \citep{barth2005,gilmore2007}.

\paragraph{Deterministic sequences.}
We condition the model on four binary sequences with known generators (\cref{fig:headline:sequences}).
Each term of Rudin--Shapiro is a simple function of the binary digits of its index, yet its correlations vanish at every lag, so no pairwise or low-order statistic predicts it \citep{allouche2003}.
The cost falls to $0.04$ bits per symbol by a million digits.
The Kolakoski sequence is its own run-length encoding \citep{kolakoski1965}: it has no closed form and no period, and each new term is read off an earlier one.
The cost falls to $0.11$ by a hundred thousand digits and stays near there, $0.13$ at a million.
On the prime indicator the cost settles at $0.28$ after one million digits, below the $0.37$ paid by a predictor that knows only how primes thin out as $1/\ln n$.
On the binary digits of $\pi$ the cost stays at one bit throughout: the sequence is computable, but nothing in a prefix reveals the rule.

\begin{table}[t]
  \centering
  \caption{%
    \textbf{PFLM compresses real data from every tested domain below an online bigram, gzip, and PPMd.}
    Bits per byte, mean over windows of $524{,}288$ bytes, up to $32$ per domain across its categories, against three baselines: an online bigram (first-order statistics), gzip (repetition), and PPMd (high-order adaptive statistics).
    \pflm{} is lowest in every row.
  }
  \label{tab:domains}
  \begin{tabular}{lrrrr}
    \toprule
    Domain & bigram & gzip & PPMd & PFLM \\
    \midrule
    Source code        & 3.76 & 2.40 & 1.82 & \textbf{1.54} \\
    Molecules          & 2.36 & 1.47 & 0.95 & \textbf{0.82} \\
    Protein            & 4.21 & 4.44 & 4.13 & \textbf{3.89} \\
    DNA                & 1.99 & 2.25 & 2.09 & \textbf{1.94} \\
    Symbolic music     & 3.59 & 3.93 & 2.99 & \textbf{2.63} \\
    Audio ($\mu$-law)  & 5.32 & 6.35 & 4.90 & \textbf{4.18} \\
    \bottomrule
  \end{tabular}
\end{table}

\paragraph{Other domains.}
We test the same weights on six further domains (\cref{tab:domains}): speech as 8-bit $\mu$-law audio \citep{panayotov2015}, piano performances as MIDI events \citep{hawthorne2019}, proteins as residue strings \citep{uniprot2023}, source code \citep{rosettacode}, a bacterial genome \citep{blattner1997}, and molecules as SMILES strings \citep{polykovskiy2020}.
We compare against three general-purpose compressors: an online bigram, gzip, and PPMd.
On every domain \pflm{} has the lowest bits per byte. The margin over the best compressor is smallest on DNA and largest on audio and music.

\section{Related Work}
\label{sec:related}

Prior-Fitted Networks train transformers on synthetic samples from a designed prior \citep{mueller2022,hollmann2023}.
The trained network performs amortized Bayesian inference on real data it has never seen.
The recipe extends to domains such as in-context regression of function classes \citep{garg2022}, survival analysis \citep{seletkov2026}, and tabular prediction at scale \citep{hollmann2025}.
Whether the recipe extends to in-context learning of natural language had not been tested.

In cognitive science, the recipe has been carried to language. \citet{mccoy2025} distill a Bayesian prior over formal languages into a network by meta-learning, giving it the inductive biases for rapid language learning, then train the network on real English.
Our prior holds no language, only generic dynamics, and \pflm{} never sees real text: what it knows of natural language, and of structure beyond it, it infers from context alone.

\emph{Learning Universal Predictors} \citet{graumoya2024} meta-trains transformers on sequences from random programs run on \acp{utm}, approximating the Solomonoff prior \citep{solomonoff1964}.
\citeauthor{graumoya2024} is \pflm{}'s closest sibling, with one substitution: a \ac{utm} prior over arbitrary programs in place of a domain-targeted \ac{scm}.
They ask whether transformers approach Solomonoff in general.
We ask whether a domain-targeted prior transfers to a specific natural domain not seen during training.

Language models have been pretrained on synthetic text before.
\emph{TinyStories} \citep{eldan2023} produces coherent English from small models trained only on GPT-generated stories using a 3-year-old's vocabulary.
\emph{Textbooks Are All You Need} \citep{gunasekar2023} extends the same recipe to code via synthetic textbooks.
Earlier work showed that pretraining on structured non-linguistic sequences (MIDI music, Java code, parentheses) transfers to real text via syntactic structure alone \citep{papadimitriou2020}.
The first two pretrain on synthetic text; the third shows that even non-linguistic structure transfers. \pflm{}'s prior shares neither content nor syntax with text, only its statistics.
The \emph{BabyLM} challenge \citep{warstadt2023} takes the opposite route, keeping the data real and capping it at a child's budget of 10M or 100M words. The cap bounds the corpus, not the exposure: winning entries revisit it for hundreds of epochs, whereas \pflm{} reads each byte of real text once.

\Ac{icl} was first demonstrated at scale in \emph{GPT-3} \citep{brown2020}: an autoregressive transformer performs new tasks from a prompt without weight updates.
It emerges only from bursty, heavy-tailed training data, not from i.i.d.\ data \citep{chan2022}, and has been formalized as implicit Bayesian inference over a latent task variable \citep{xie2022}.
\pflm{} is a clean test case for these accounts: its entire in-context ability comes from a prior of known structure, and it is evaluated on a domain its training data does not contain.

\section{Conclusion}
\label{sec:conclusion}

We presented \pflm{}, a byte-level transformer trained on no real data, which learns to predict language, numerals, and deterministic sequences it never saw during training.
They differ as data but pose the same inference problem: identify the structured source behind a prefix and predict its continuation. Trained by next-token prediction on samples from a prior over such sources, the model approximates the Bayesian predictor for that problem (\cref{sec:method:inference}).
On Wikipedia text in six languages, bits per byte fall steadily as the context grows with the weights fixed. Beyond language, \pflm{} learns to count through carries, to compare magnitudes, and to add approximately. It predicts Rudin--Shapiro, Kolakoski, and the prime indicator far below chance. The model has not learned a language. It has learned to learn one.

Context length currently limits how far in-context learning can get. The trained context is one million bytes, about two hundred thousand English words, and at that length the prediction on Wikipedia still falls short of what modern language models reach after training on trillions of tokens. We expect two lines of future work to be the most rewarding: applying the prior to domains with little data, such as low-resource languages, undeciphered scripts, or animal communication, and joint pretraining on real text and prior samples to enhance the in-context learning of conventional language models.

\clearpage
\bibliographystyle{plainnat}
\bibliography{references}

\clearpage
\appendix
\section{Prior configuration}
\label{app:prior-config}

\Cref{tab:prior-config} lists the sampled ranges of every component of the prior (\cref{sec:method:prior}).
A range is drawn once per language where the table says so and once per node otherwise.

\begin{table}[h]
  \centering
  \caption{Sampled ranges of the prior. Per-language draws fix a value for the whole language; per-node draws are made independently at each node.}
  \label{tab:prior-config}
  \begin{tabular}{llp{0.42\linewidth}}
    \toprule
    Component & Drawn & Range \\
    \midrule
    Nodes $d$                         & per language & uniform in $[8, 32]$ \\
    Observed coordinates $d_{\mathrm{obs}}$ & fixed   & $8$ (with $L = 2$ levels: $V = 256$) \\
    Parent-set size $|\pa(i)|$        & per node     & uniform in $[2, 8]$, capped at $d - 1$ \\
    Update layers $K$                 & per language & uniform in $\{1, 2\}$ \\
    Equation family $f_i$             & per node     & uniform over the six families of \cref{tab:equations} \\
    Coefficient scale                 & per language & weights $\mathcal{N}(0, s_w^2)$ with $s_w \sim U(1, 2.5)$, biases $\mathcal{N}(0, s_b^2)$ with $s_b \sim U(0.1, 0.5)$; weight scale divided by $\sqrt{\text{fan-in}}$ \\
    Activation $\sigma_i$             & per node     & uniform over $\{\tanh, \mathrm{softsign}, \clip, \sin, \mathrm{id}\}$ \\
    Noise scale $\eta$                & per language & uniform in $[0, 1/20]$ \\
    Firing rate $r_i$                 & per node     & powers of two, most mass on $1$, geometric tail to $2^6$ \\
    Phase $\phi_i$                    & per node     & uniform in $\{0, \ldots, r_i - 1\}$ \\
    Initial state $\mathbf{h}_0$      & per sequence & one of six families on $[-1, 1]$: uniform on a sub-interval, Gaussian, Beta, two-point near $\pm 1$, bimodal, sparse \\
    Rollout length $T$                & fixed        & the training context length \\
    \bottomrule
  \end{tabular}
\end{table}

\section{Model and training}
\label{app:training}

\Cref{tab:model-config} gives the shape of the model of \cref{sec:method:model}, and \cref{tab:train-hparams} the optimization.

\begin{table}[h]
  \centering
  \caption{Model configuration: blocks of three gated linear-attention layers followed by one sliding-window attention layer, SwiGLU MLPs, and the 256-byte vocabulary. The parameter count is measured.}
  \label{tab:model-config}
  \begin{tabular}{lr}
    \toprule
    Parameters                       & 295M         \\
    Layers (linear attn.\ + SWA)     & 20 (15 + 5)  \\
    Hidden size                      & 1024         \\
    MLP intermediate size            & 2048         \\
    Attention heads / KV heads       & 16 / 8       \\
    Attention head dim               & 128          \\
    Linear-attn.\ value / key heads  & 16 / 8       \\
    Linear-attn.\ head dim (key, value) & 128       \\
    Vocabulary                       & 256          \\
    \bottomrule
  \end{tabular}
\end{table}

\pflm{} is trained in six stages of increasing context length, each stage initialized from the last: $2^{10}$, $2^{12}$, $2^{14}$, $2^{16}$, $2^{18}$, and $2^{20}$ bytes, for $20$, $60$, $40$, $15$, $10$, and $5$ billion tokens respectively, $150$ billion in all.
Every stage draws its sequences afresh from the prior at that length, so no sequence is seen twice, and the learning rates are set per stage.

\begin{table}[h]
  \centering
  \caption{Optimization hyperparameters.}
  \label{tab:train-hparams}
  \begin{tabular}{lp{0.6\linewidth}}
    \toprule
    Hyperparameter & Value \\
    \midrule
    Optimizer               & NorMuon~\citep{li2025} for matrix parameters; AdamW~\citep{loshchilov2019} for embeddings, head, and norms \\
    Matrix learning rate    & $8\times10^{-4}$ \\
    AdamW learning rate     & $1.6\times10^{-5}$ \\
    Schedule                & warmup--stable--decay~\citep{hu2024}, 500 warmup steps \\
    Tokens per stage        & 20B, 60B, 40B, 15B, 10B, 5B at $2^{10}$ to $2^{20}$ bytes of context \\
    Sliding window          & $2^{18}$ bytes \\
    $z$-loss weight         & $1\times10^{-4}$ \\
    Hardware                & $4\times$ NVIDIA H100 \\
    \bottomrule
  \end{tabular}
\end{table}

The gated linear-attention layers run on the FlashQLA kernels \citep{flashqla2026}, built on \texttt{flash-linear-attention} \citep{fla2024}; the sliding-window attention layers use FlashAttention-3 \citep{shah2024}.

\section{Byte-frequency profiles}
\label{app:byte-frequency}

\begin{figure}[h]
  \centering
  \includegraphics[width=\linewidth]{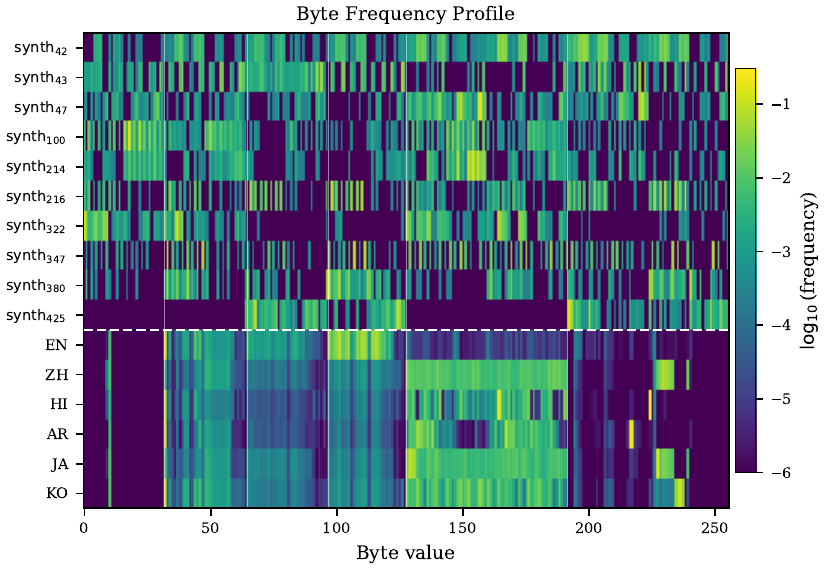}
  \caption{%
    \textbf{The prior produces samples whose byte distributions span the breadth of natural language without sharing any specific script.}
    Byte-frequency profile (log scale) for ten random samples from the prior (above the dashed line) and the six natural-language corpora (below).
    Each row is one source's distribution over byte values 0--255.
    Synth samples each use a substantial subset of the byte vocabulary, and the active subset varies from sample to sample.
    Each natural language pins to a characteristic UTF-8 byte range fixed by its script.%
  }
  \label{fig:byte_frequency}
\end{figure}

\Cref{fig:byte_frequency} places ten random samples from the prior of \cref{sec:signatures} beside the six natural-language corpora.
Each natural language pins to the script-specific UTF-8 range its writing system occupies; each prior sample spreads its mass over a different region of the byte space, never the same one twice.
The two cover the same breadth without sharing a script.

\section{Evaluation protocol}
\label{app:eval}

Every evaluation of \cref{sec:exp:language,sec:exp:beyond} uses the \pflm{} checkpoint with frozen weights; the model adapts to a corpus only through its context, never through a gradient step.

\paragraph{Encoding.}
Each corpus is a stream of raw bytes, and the model's vocabulary is the $256$ byte values throughout (\cref{sec:method:model}).
Natural-language text is encoded with UTF-8; the binary digit sequences map each symbol to one of two byte values.
A ``byte'' on every context axis is thus one encoded byte, and bits per byte (BPB) is measured per byte rather than per character.
This matters most for the multilingual panel of \cref{fig:headline}: under UTF-8 the Latin script of English costs about one byte per character, Arabic about two, and Devanagari, Chinese, Japanese, and Korean about three, so a curve's level reflects the encoding as well as the language.

\paragraph{Natural language.}
The six languages are English, Chinese, Hindi, Arabic, Japanese, and Korean, taken from the \texttt{wikimedia/wikipedia} \texttt{20231101} Wikipedia dump (one date-prefixed configuration per language).
Articles are streamed, concatenated, and cut into fixed-length byte sequences held out from training; the statistical-signature analysis of \cref{sec:signatures} uses $500$ sequences of $4096$ bytes per language.
The acquisition curves of \cref{fig:headline} condition on $32$ sequences of $1{,}048{,}576$ bytes per language, each starting at an article boundary, and report the mean cost at every position with the standard error across sequences.

\paragraph{Metric and bands.}
BPB at position $t$ is $-\log_2 p_\theta(y_t \mid y_{<t})$, the model's surprisal at the next byte given the prefix.
The acquisition curves of \cref{fig:headline} plot this against $t$ as the prefix grows, with the weights fixed throughout.
Uncertainty bands are defined in each figure's caption.

\paragraph{Digit sequences.}
The four sequences of \cref{fig:headline:sequences} are generated deterministically from their definitions: the Rudin--Shapiro sequence (parity of the number of pairs of adjacent ones in the binary expansion), the Kolakoski sequence (its own run-length encoding), the prime indicator $\chi(n)=\mathbf{1}[n\ \text{prime}]$, and the binary digits of $\pi$.
Each is over a binary alphabet mapped to two byte values, so chance is one bit per symbol.
The floor quoted for the prime indicator is the cost of a predictor that knows only the density of primes near $n$, the binary entropy of $1/\ln n$, which is $0.37$ bits at $n = 10^6$.

\paragraph{Number.}
Counting reads the stream $2, 3, 4, \ldots$ up to $99{,}999$ as decimal numerals separated by commas and scores each numeral by exact match of the greedy decode given the stream so far; numbers are grouped by digit count and by whether the increment carries.
Comparison and approximate addition are few-shot: each trial is $k$ example lines followed by a query line, scored from the logits at the answer position restricted to the two label bytes.
A comparison line reads $n_1,n_2{=}L$ with $L=1$ when the second numeral is larger; both numerals have three digits, and query pairs are drawn per ratio bin of the smaller to the larger and never appear among the examples.
An approximate-addition line reads $n_1{+}n_2,n_3{=}L$ with $L=1$ when the sum exceeds $n_3$; $n_3$ exceeds either addend alone, and queries are binned by the ratio of the sum to $n_3$.
Each cell has $500$ queries, at $k = 128$, $1024$, and $8192$.

\end{document}